\documentclass{article}

\usepackage{arxiv}

\usepackage[utf8]{inputenc} % allow utf-8 input
\usepackage[T1]{fontenc}    % use 8-bit T1 fonts
\usepackage{hyperref}       % hyperlinks
\usepackage{url}            % simple URL typesetting
\usepackage{booktabs}       % professional-quality tables
\usepackage{amsfonts}       % blackboard math symbols
\usepackage{nicefrac}       % compact symbols for 1/2, etc.
\usepackage{microtype}      % microtypography
\usepackage{lipsum}
\usepackage{graphicx}

\title{A Novel \emph{Approximate Hamming Weight} Computing for Spiking Neural Networks: an FPGA Friendly Architecture}

\author{
  Kaveh Akbarzadeh-Sherbaf \\
  \texttt{k.akbarzadeh@ut.ac.ir} \\
  \And
  Mikaeel Bahmani \\
  \texttt{mikaeel.bahmani@ut.ac.ir} \\
  
  \And
  Danial Ghiaseddin \\
  \texttt{danial.ghiaseddin@ut.ac.ir} \\
  
  \And
  Saeed Safari \\
  \texttt{saeed@ut.ac.ir} \\
  
  \And
  Abdol-Hossein Vahabie \\
  \texttt{vahabi@ipm.ir} \\
  
}
\begin{document}
\maketitle

\begin{abstract}
 Hamming weights of sparse and long binary vectors are important modules in many scientific applications, particularly in spiking neural networks that are of our interest. To improve both area and latency of their FPGA implementations, we propose a method inspired from synaptic transmission failure for exploiting FPGA lookup tables to compress long input vectors. To evaluate the effectiveness of this approach, we count the number of `1's of the compressed vector using a simple linear adder. We classify the compressors into shallow ones with up to two levels of lookup tables and deep ones with more than two levels. The architecture generated by this approach shows up to 82\% and 35\% reductions for different configurations of shallow compressors in area and latency respectively. Moreover, our simulation results show that calculating the Hamming weight of a 1024-bit vector of a spiking neural network by the use of only deep compressors preserves the chaotic behavior of the network while slightly impacts on the learning performance. 
\end{abstract}


\section{Introduction}
\label{sec:intro}
The spiking neural networks (SNNs) are the possible power-efficient alternatives to the power-hungry artificial neural networks (ANNs). The human Brain, the source of inspiration for the SNNs, consumes much less power in comparison to the state-of-the-art hardware accelerators of conventional ANNs. To break this power wall, researchers strive to mimic the Brain architecture by introducing novel spiking hardwares \cite{Merolla2014b, Davies2018} and algorithms \cite{Jaeger2001, Maass2002, Sussillo2009, Nicola2017, Gilra2017}. Since Brain networks operate between order and chaos \cite{Gao2016}, the most recent bio-inspired learning algorithms employ chaotic SNNs as their initial untrained networks \cite{Sussillo2009, Nicola2017, Gilra2017}. 

In an SNN, a presynaptic action potential, also known as a spike, produces a PSP on the membranes of postsynaptic neurons. The accumulation of charge particles on the membrane of the  postsynaptic neuron may lead the postsynaptic neuron to fire an action potential. In a typical SNN, each neuron may receives inputs from thousands of synapses. To accumulate the PSPs, in a binary network, the number of received spikes should be counted \cite{Kaveh2018}. 

According to the notation of \cite{Kaveh2018}, the connectivity pattern of a biological neuronal network is usually depicted as an $N\times N$ sparse matrix with the density below 20\% \cite{Seyed-allaei2015}. Each row (or column) of this matrix shows the presynaptic (or postsynaptic) neurons of a given neuron. Here we call this N-element vector connection vector (CV) and show its density with $d_{CV}$. Furthermore, it is common to store the firing activity of the latest snapshot in another N-element vector called spike vector (SV). We use a similar notation ($d_{SV}$) to show the density of spike vector. Applying the bitwise AND between SV and CV produce a new N-element vector in which each `1' represents a presynaptic neuron fired in the previous time step. The density of the resultant vector (RV) is $d_{CV}\times d_{SV}$. In an extreme case which SV is an all-ones vector, RV is still sparse ($d_{RV}<20\%$). We certainly do not expect this extreme case and SV is usually a sparse vector too. Hence RV is a much sparser vector ($d_{RV}<<20\%$). Hamming weight of RV is the number of its nonzero elements and it equals the number of received spikes. Therefore, Hamming weight of RV represents the synaptic efficacy.

%According to the notation of \cite{Kaveh2018}, for a network with N neurons, the arrived spikes to a neuron are the AND function of two N-bit binary vectors called CV and SV. The vector PV is the connection vector of that neuron and illustrates its presynaptic neurons and the vector SV is a vector that indices of its nonzero elements represent the neurons fired an action potential. Hrer

In the Brain, only 10-30\% of presynaptic spikes produce a postsynaptic potential (PSP) due to the synaptic transmission failure \cite{Hessler1993, Markram1996}. In this paper, for the first time, we implement an approximate Hamming weight calculator inspired by this phenomenon in hardware. Undoubtedly, the approximation should not change the activity regimes of chaotic SNNs.

The remainder of the paper is organized in three sections. In the next section we review some related works and present our method. Section \ref{sec2:proposed architecture} shows the details of our proposed architecture. In section \ref{sec3}, we compare the accuracy, resource usage, and delay of the proposed architecture with its exact counterpart. Finally, in section \ref{sec4}, we conclude the paper with a short but informative discussion on the obtained results.

\section{Backgrounds and goals}
\label{sec1:backgrounds and goals}

Several applications in coding theory, cryptography, and neural network disciplines require counting the `1's' population in an efficient way. Its wide range of applications encouraged Intel to support POPCNT (population count) instruction beginning with the Nehalem architecture \cite{Kurzak2010}. 

Researchers have proposed efficient techniques for the accurate Hamming weight computation in both software and hardware frameworks. A good review of software implementations including the ones who had exposed scalar and vector parallelism was conducted in \cite{Morancho2014}. Moreover, it presented a new hybrid scalar-vector implementation, which had improved its best prior software realizations up to 1.23$\times$. On the other hand, new counting-based methods for comparing either two Hamming weights or a Hamming weight with a fixed threshold were presented in \cite{Parhami2009}. Furthermore, it reviewed the arithmetic architectures, which had been proposed to lower the delay and cost complexity. There is a growing interest in implementing complex computational systems using reconfigurable devices, particularly field programmable gate arrays (FPGAs). The work presented in \cite{Sklyarov2015a} is among the ones concentrating on FPGA implementations of Hamming weights. They employed a multi-level combinational circuit to calculate Hamming weights of $2^L-bit$ at level $L$. Each level uses the previous level output to reduce the original input step by step. It finally results in a decimal vector that indicates the number of `1's in the input vector. Each level of their proposed architecture well matches the FPGA lookup tables (LUT). Unfortunately, despite all these efforts, the hardware resources and consequently the latencies of the circuits increase dramatically with the number of input bits.
\begin{figure}[tbp]
    \centering
	\includegraphics{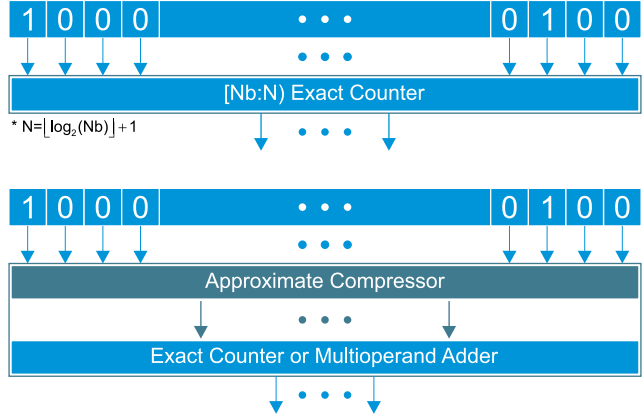}
	\caption{\textbf{Top.} An exact counter is a traditional method to count the number of `1's entries of an input vector. \textbf{Bottom.} Reducing the number of inputs to the exact counter using an approximate compressor.}
	\label{fig:exactCompressor}
\end{figure}

%\begin{figure}
%	\includegraphics[width=80mm]{figs/fig2}
%	\caption{Reducing the number of inputs to the exact counter using an approximate compressor.}
%	\label{fig:approxCompressor}
%\end{figure}
While above-mentioned papers include applications such as rank order filters that require the exact Hamming weight \cite{Pedroni2004}, there are applications, which give robust performances in the presence of the inaccuracy. These types of applications give us a chance of reducing both hardware cost and latency. Among such applications, we are particularly interested in SNNs where we need to count roughly the number of spikes arrived at synaptic clefts \cite{Kaveh2018}. The synaptic transmission failure is a biological evidence that shows an error up to almost 90\% is tolerable in counting the number of spikes (Hamming weight of RV). Finally, Postsynaptic neurons use these approximate numbers to quantify synaptic efficacies.

To calculate the Hamming weight using the conventional exact methods introduced in \cite{Swartzlander1973, Parhami2009, Sklyarov2015a}, an exact compressor is required (Fig.~\ref{fig:exactCompressor} Top)\footnote{Throughout the paper, [N:n) denotes a circuit that counts the number of `1's in its N-bit input string and stores the result in an n-bit output string in either an exact or an approximate manner.}. To lower the hardware cost and latency, for the cases with a large number of input bits, here we propose a set of multi-level approximate compressors, which are used to reduce the number of inputs to the exact compressor (Fig.~\ref{fig:exactCompressor} Bottom). It, therefore, results in an affordable circuit that shows a trade-off between its hardware requirements, e.g., resource usage and latency, and the accuracy.

\begin{table}[tbp]
    \centering
	\caption{One-level and two-level approximate compressors with their corresponding percentages of correct outputs from all possible input vectors.}
	\label{tab:tab1}
	\begin{tabular}{clllc}
		\hline
		Config.	&Level 1				&Level 2				&Overall				&Correct outputs (\%)\\
		\hline
		(A) &[6:1)					&-						&[6:1)					& $\approx 11$ \\
		(B) &[5:2$^{\diamondsuit}$)	&-						&[5:2$^{\diamondsuit}$)	& $50$ \\
		(C) &[5:2)					&-						&[5:2)					& $\approx 81$ \\
		(D) &[6:1)					&[6:1)					&[36:1)					& $\approx 5.4e-08$ \\
		(E)	&[6:1)					&[5:2$^{\diamondsuit}$)	&[30:2$^{\diamondsuit}$)& $\approx 3.8e-06$ \\
		(F)	&[6:1)					&[5:2)					&[30:2)					& $\approx 4.7e-06$ \\
		(G)	&[5:2$^{\diamondsuit}$)	&[6:1)					&[15:1)					& $\approx 4.9e-02$ \\
		(H)	&[5:2$^{\diamondsuit}$)	&[5:2$^{\diamondsuit}$)	&[25:4$^{\diamondsuit}$)& $\approx 1.9e-02$ \\
		(I)	&[5:2$^{\diamondsuit}$)	&[5:2)					&[25:4)					& $\approx 1.9e-02$ \\
		(J)	&[5:2)					&[6:1)					&[15:1)					& $\approx 4.9e-02$ \\
		(K)	&[5:2)					&[5:2$^{\diamondsuit}$)	&[25:4$^{\diamondsuit}$)& $\approx 1.7e-03$ \\
		(L)	&[5:2)					&[5:2)					&[25:4)					& $\approx 7.8e-03$ \\
		\hline
	\end{tabular}
\end{table}

\section{Proposed Architecture}
\label{sec2:proposed architecture}
FPGA LUTs in general terms are designed to implement various binary functions. They, therefore, should be fabricated in such a way that meets diverse requirements. It seems that small lookup tables distributed throughout the device have the best design trade-offs. To build larger memory, it is possible to cascade multiple LUTs. Such a design opens up an opportunity to employ LUTs in the core of an approximate hardware, especially for our domain of interest. Let $I$ and $O$ be the number of LUTs' inputs and outputs respectively, then $\lceil\log_{2}(I)\rceil>O$ for a given LUT, particularly in 7-series Xilinx FPGAs, which are our target family devices. This expression means that a single LUT cannot count the number of `1's in the input vector when the number of `1's are greater than $2^O$. Indeed, it is an approximate counter that can show a near exact performance for the sparse inputs. 
\subsection{FPGA LUTs as One-Level Approximate Compressors}
\begin{figure}[tbp]
    \centering
	\includegraphics[width=0.5\textwidth]{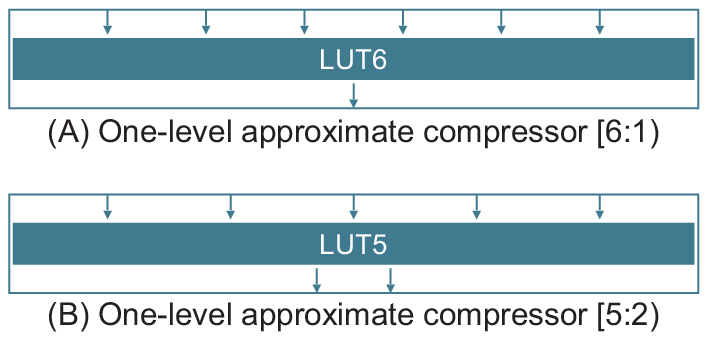}
	\caption{Overview of the lookup tables of Xilinx 7 series FPGAs. They could be used as one-level compressors. }
	\label{fig:oneLevel}
\end{figure}

LUTs of Xilinx 7-series FPGAs can be configured as either a six-input (LUT6) or two five-input (LUT5) boolean function generators (Fig. \ref{fig:oneLevel}). Since LUT6 has only one output, it could count up to only one nonzero element. It, therefore, is an approximate counter that gives an accurate result just in almost $(7/64)\times 100 \approx 11\%$ of all possible cases (Configuration A in Table \ref{tab:tab1}). Despite its great error at first glance, it could produce more accurate results under sparse conditions, which will be shown in section \ref{sec4}. From now on, [6,1) stands for a one-level approximate compressor built by LUT6 (Fig. \ref{fig:oneLevel}A). Although this compressor contains inaccuracy, it is worth considering due savings made in resource consumption. It compresses an input vector into a six times smaller vector that is the input to the second level exact compressor illustrated in Fig. \ref{fig:exactCompressor}. Reducing the number of inputs of an exact compressor brings its resource usage down exponentially.
\begin{figure}[tbp]
    \centering
	\includegraphics[width=0.6\textwidth]{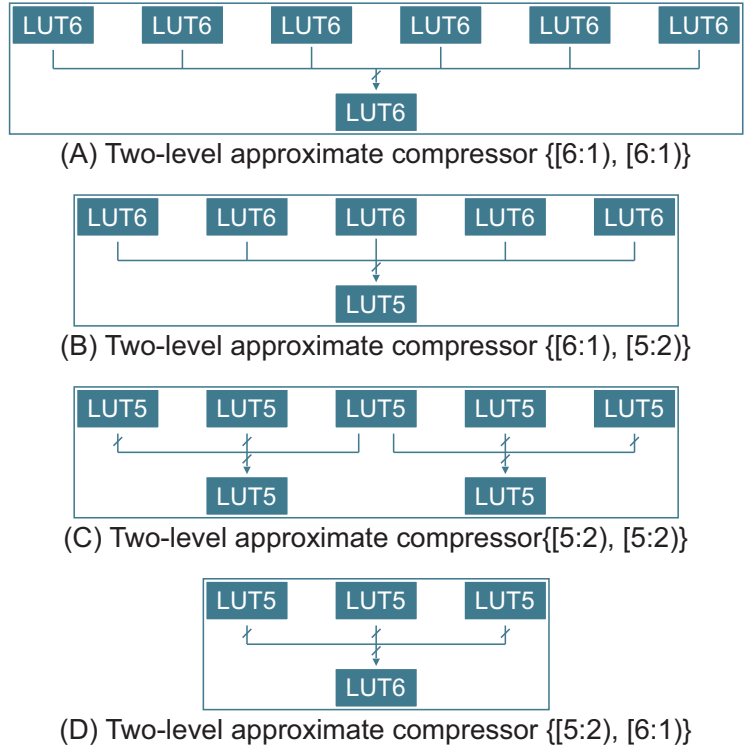}
	\caption{Various configurations of two-level approximate compressors using FPGA LUTs.}
	\label{fig:twoLevel}	
\end{figure}

LUT5, on the other hand, has two output bits along with one less input. Thus, it results in a more accurate and less compressed vector. We interpret two output bits as a binary code in two ways: (1) a positional and (2) a non-positional number. Throughout the paper, [5:2) and [5:2$^{\diamondsuit}$) denote the positional and nonpositional numbers respectively. The outputs of both five-input compressors are $O=\{``00",``01",``10",``11"\}$. In a [5:2) compressor,  each object of the set $O$ is a binary number that shows 0, 1, 2, or 3 ones of the input vector respectively. Hence, it could count up to three ones of the input vector. On the contrary, the number of objects' ones of the set $O$ determines the number of input bits with values of `1's for the [5:2$^{\diamondsuit}$) compressor. Although this coding scheme has the higher error, it hopefully lowers the resource usage of the second level exact compressor mainly because the exact compressor is a multi-operand adder with one-bit operands (aka counter) and two-bit operands for the [5:2$^{\diamondsuit}$) and [5:2) cases respectively.

The [5:2$^{\diamondsuit}$) and [5:2) compressors give accurate results in almost $(16/32)\times 100 =50\%$ and $(26/32)\times 100 \approx 81\%$ of all possible cases respectively (Configuration B and C shown in Table \ref{tab:tab1}). As it will be shown later, they could show much better results in sparse conditions. While these compressors lead to more accurate results, they consume more resources than [6:1). A five-input compressor converts the input vector to a 2.5 times smaller vector, in contrast to six-input compressor, which produces six-times smaller output vector.

\begin{figure*}[tbp]
    \centering
	\includegraphics[width=\textwidth]{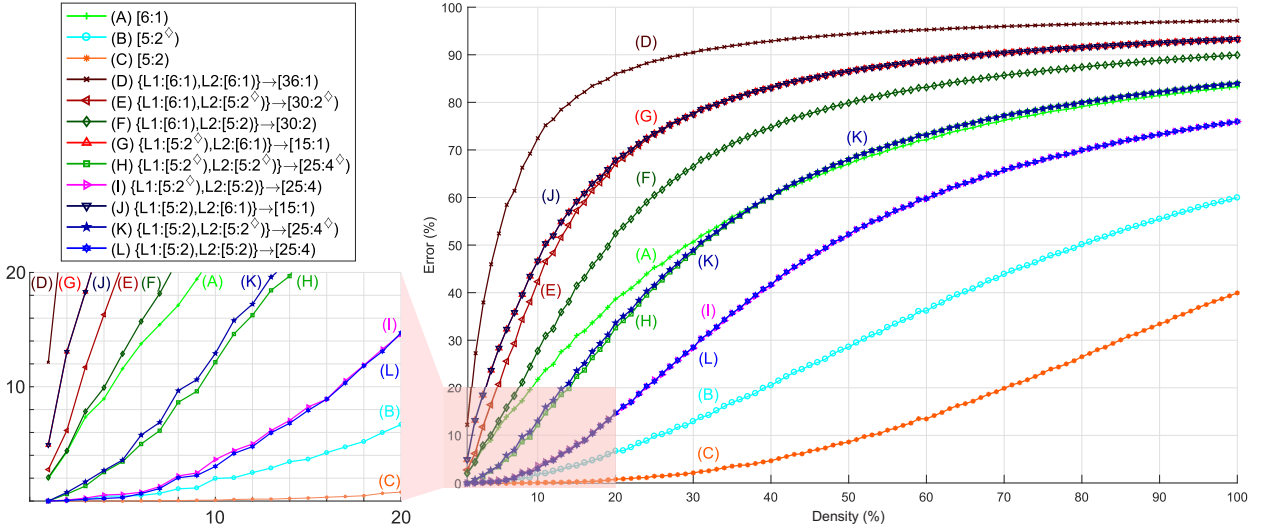}
	\caption{Errors of one-level and two-level compressors for different densities.}
	\label{fig:errorChart}
\end{figure*}

\subsection{Two-Level Approximate Compressors}
To further compress the vector injected to the exact level, it is possible to cascade the LUT6 ([6:1) compressor) and LUT5s ([5:2$^{\diamondsuit}$) and [5:2) compressors) in multiple levels. We thus obtain four main categories for the case of two-level approximate compressors, which is of our interest in this paper (Fig. \ref{fig:twoLevel}). Since the LUT5 has two output coding styles in this paper, there are nine distinct configurations for two-level compressors. Fig. \ref{fig:twoLevel}A illustrates a configuration that uses LUT6s to form a [36:1) compressor. Fig. \ref{fig:twoLevel}B shows the [30:2) and [30:2$^{\diamondsuit}$) compressors that their outputs are 15 times smaller than their inputs in term of the number of bits. Four of twelve possible configurations are stated in Fig. \ref{fig:twoLevel}C, which includes LUT5s.  These four configurations consists of two [25:4) and two [25:4$^{\diamondsuit}$) compressors (see Table \ref{tab:tab1} for more details about the arrangements of the LUTs). The left configurations are two types of [15:1) compressors represented in Fig. \ref{fig:twoLevel}D where three LUT5s and an LUT6 are placed in the first and second level respectively.

Table \ref{tab:tab1} shows the percentage of exact outputs for 12 configurations. According to this table, the [5:2) one-level compressor have significant accuracy, and it could be a great compressor in a Hamming weight counter. As it will be said later, this compressor is the best one from the accuracy point of view. On the other hand, although the percentage of correct outputs of the [6:1) one-level compressor is orders of magnitude higher than two-level compressors, it does not guarantee it will have higher total accuracy than two-level ones. This is due to the fact that more than one bit with value 1 is not recognizable in the only six-bit distance. It will be shown that the [6:1) compressor in any architecture highly reduces the accuracy of the circuit. 

\subsection{Deep Compressors}
\label{sec:deep}
The density of RV (see Section \ref{sec:intro}) are usually several orders of magnitude less than one. Consequently, we can employ deeper compressors with more levels of LUT6s and LUT5s instead of shallow one-level and two-level ones. To show the possibility of replacing the shallow compressors with deep ones, we will examine their impact on the SNN performance in Section \ref{sec:impa}. There, the [216:1), [540:1), and [1024:1) compressors, along with shallow [6:1) and [36:1) compressors, will be applied to a network of 1024 Izhikevich neurons.

\subsection{Final Exact Counter}
Although the approximate compressors reduce the number of input vector bits, it is finally necessary to calculate the Hamming weight of the reduced input vector using an exact adder. There are many techniques to implement such a circuit. One implementation could be a traditional tree or linear adder \cite{Parhami2010}. On the other hand, it is possible to employ dedicated architectures introduced for Hamming weight calculation \cite{Parhami2009}. In this paper, we want to examine the effects of approximate counters on the delay and resource usage. Indeed approximate counter is what matters---the type of the exact counter is of secondary importance. Therefore, we employ a simple linear adder to show the efficacy of the proposed compressors.
\section{Results and Discussion}
\label{sec3}
In this section, we estimate the error of our proposed architecture under different sparsities. We also show the impact of the synaptic transmission failure modeled by the proposed architecture on the chaotic activity of a chaotic SNN. Moreover, we use the Xilinx Vivado design suite to synthesize the architecture in order to report the latency and area cost.

\subsection{Accuracy of the Architecture}
To evaluate the accuracy, we write a OCTAVE code to simulate the shallow one-level and two-level compressors for 1024-bit inputs. In this code, for the final exact adder, we employ OCTAVE built-in "sum" function. We then inject 1000 different randomly generated binary vectors  (uniformly distributed) with a given sparsity to the model. It results in 1000 different errors for an architecture per a given sparsity. The mean and standard deviation of the obtained errors are reported in Fig. \ref{fig:errorChart}. To explore the effects of sparsity, we vary the densities of the random input vectors from 1\% to 100\%. Note that the standard deviations are far too small to be visible in this chart.

As it can be clearly seen in Fig. \ref{fig:errorChart}, while the [5:2) configuration shows the lowest error for all sparsities, the [36:1) denotes the highest errors, as we have expected. One of the interesting point to note here is the ability of even [36:1) compressor to show satisfactory errors for some applications. For example, the error of about 12\% at a density of 1\% is good enough for a few applications in spiking neural networks \cite{Kaveh2018}. Comparison of one-levels and two-levels compressors show that some two-levels' compressors (K, H, I, L) have lower errors than configuration (A) that is a one-level compressor. Furthermore, (I) and (L) compressors surprisingly present an error near (B) and (C) for densities below 10\%.

\subsection{Impact on Chaotic activity}
\label{sec:impa}

\begin{table}[tbp]
    \centering
	\caption{Synaptic failures percentages, mean firing rates, and mean squared errors of different configurations.}
	\label{tab:tab2}
	\begin{tabular}{cccc}
		\hline
		Config.	&Synaptic Failure (\%)		&MFR	&MSE			\\
		&max, mean& spikes/sec & \\
		\hline
		Exact						&0, 0		&7.30	&3.77$\times$10\textsuperscript{-3}\\
		{[6:1)} 	&1.85, 0.08	&7.26	&1.30$\times$10\textsuperscript{-3}\\
		{[36:1)} 	&2.54, 1.08	&7.33	&2.27$\times$10\textsuperscript{-3}\\
		{[216:1)} 	&9.01, 1.33	&7.28	&2.62$\times$10\textsuperscript{-3}\\
		{[540:1)} 	&17.45, 1.52	&7.29		&2.32$\times$10\textsuperscript{-3}\\		
		{[1024:1)} 	&31.29, 1.68	&7.22		&13.89$\times$10\textsuperscript{-3}\\		
		\hline
	\end{tabular}
\end{table}
\begin{figure}[tbp]
    \centering
	\includegraphics[width=0.5\textwidth]{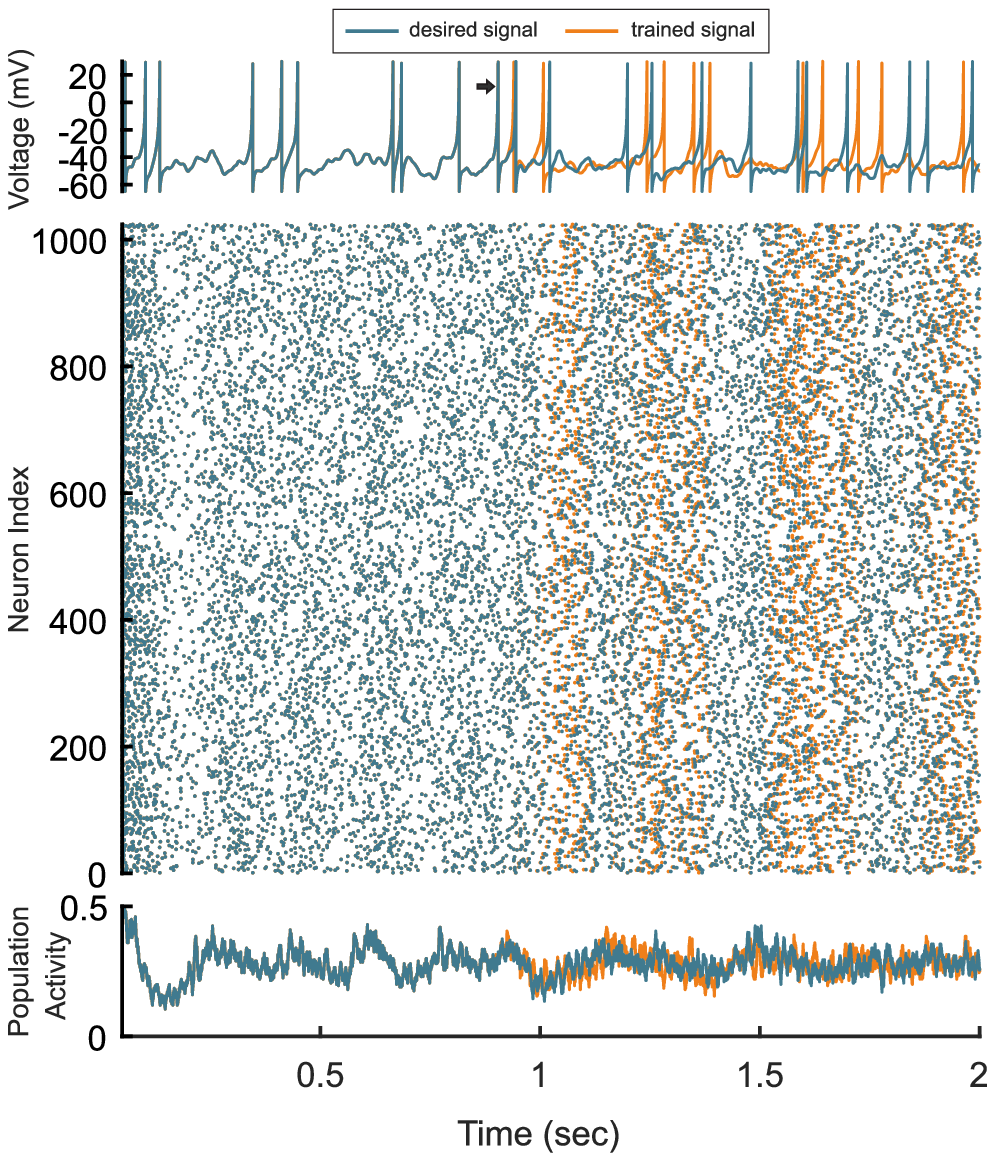}
	\caption{\textbf{Top.} Membrane voltage of the first neuron. \textbf{Middle.} Raster diagram of the network. \textbf{Bottom.} Population activity calculated by the use of moving average method with window length of 8 ms.}
	\label{fig:raster}	
\end{figure}

To test the impact of synaptic failure modeled by the proposed compressors, we choose a network of Izhikevich neurons suggested by \cite{Nicola2017} to implement a chaotic SNN. They trained their proposed network to learn chaotic attractors. This network is a 1024-neuron network with density of 10\% and thus sparsity of 90\%. The network has equal number of excitatory and inhibitory neurons, and therefore connection matrices densities of both types of neurons are 5\%. We made two modifications to the original network. First, since our proposed method is applicable to binary networks, we changed the real-valued connection matrix to a bi-valued one. Second, we replaced the exact adder used to accumulate the presynaptic spikes with five approximate ones with five different shallow and deep compressors. The first three compressors consist of one, two, and three levels of LUT6s to make [6:1), [36:1), and [216:1) compressors. Next, [540:1) compressor which can be made by [36:1) and [15:1) compressors is the main compressor. Finally, our compressor is [1024,1) compressor that can represent any [x:1) compressors with x$>$1024. To investigate the impact of proposed compressors on the network, we consider both chaotic activity and the learning performance.

We run the network for five seconds split into three intervals, namely initialization, train, and generate. In initialization interval lasted two seconds, the network run without applying target sine signal. Before evaluating the compressors' impacts, we study the exact network activity. Fig. \ref{fig:raster}a shows the membrane voltage of the first neuron for the initialization period. Moreover, the whole network activity is represented by the use of raster diagram and the population activity illustrated in Fig. \ref{fig:raster}b and Fig. \ref{fig:raster}c respectively. To show that the network is in chaos, we delete the one spike of the first neuron, indicated by a black arrow. As it is expected, the membrane voltage and the network activity deviate from the original network without the deletion of a spike. This is a clear indication of chaotic behavior \cite{Nicola2017}.
\begin{figure}[tbp]
    \centering
	\includegraphics[width=0.5\textwidth]{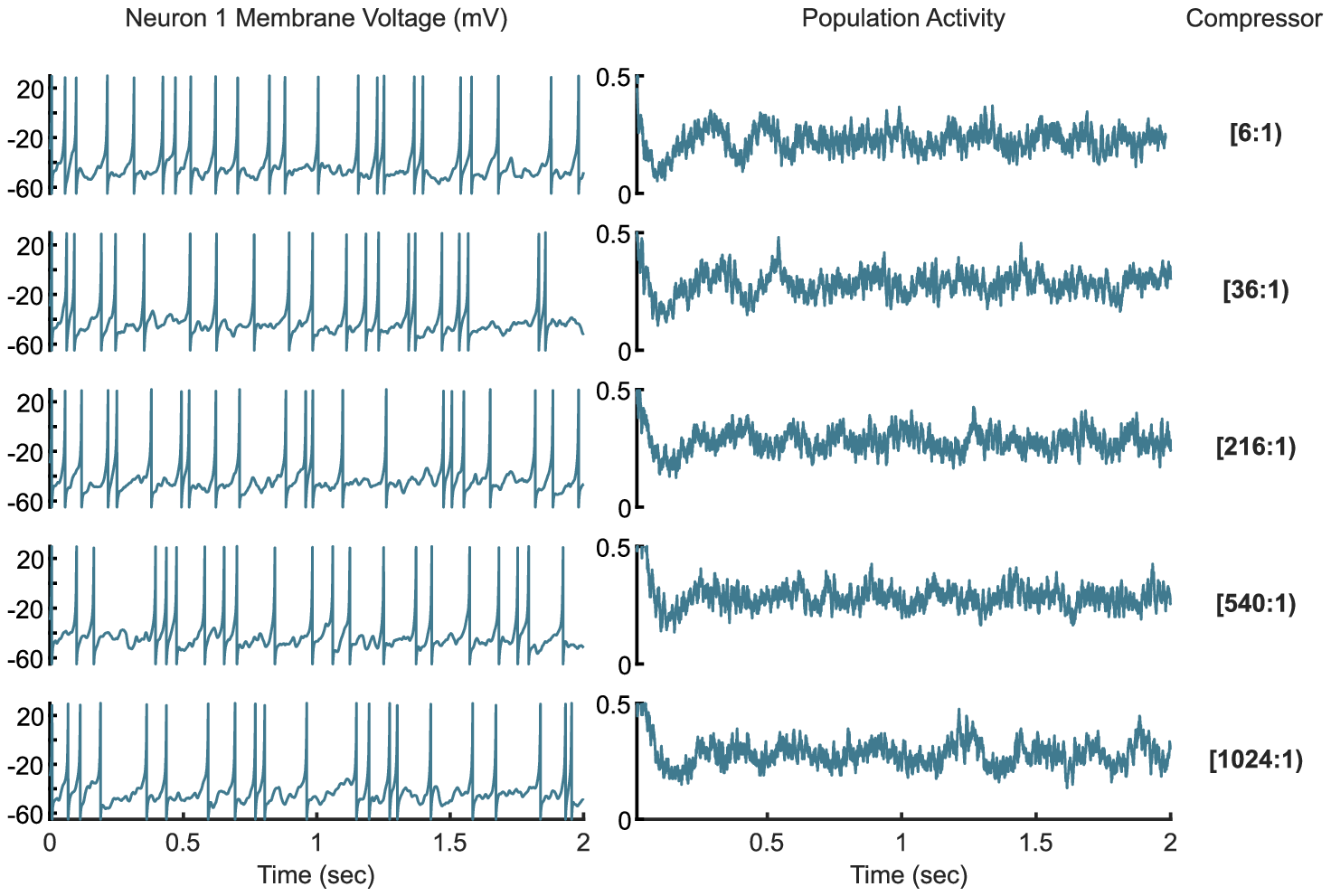}
	\caption{\textbf{Left.} Membrane voltage of the first neuron for the networks equipped with shallow and deep compressors. \textbf{Right.} The corresponding population activities.}
	\label{fig:popactivity}	
\end{figure}
\begin{figure}[tbp]
    \centering
	\includegraphics[width=0.5\textwidth]{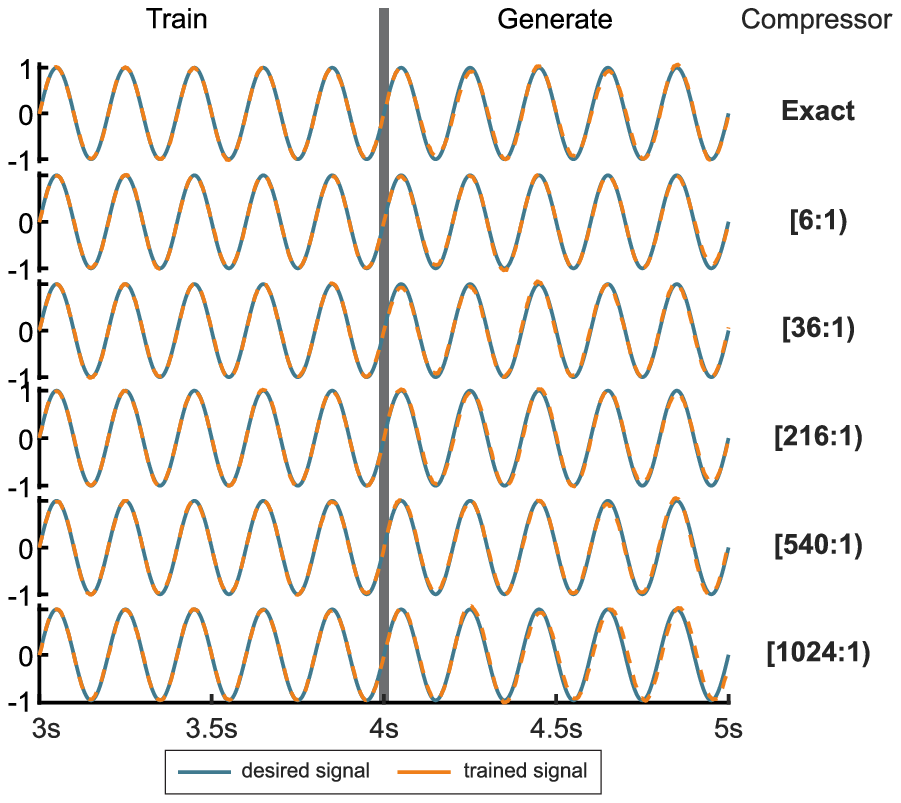}
	\caption{A 5 Hz sine wave is used as the target signal. After training for two seconds, the network runs freely to regenerate its learned dynamics.}
	\label{fig:sine}	
\end{figure}

Fig. \ref{fig:popactivity} shows the first neuron membrane voltage and population activity for five network setups with approximate Hamming weight calculators. The use of each compressor results in different membrane voltages and population activity. However, they follow similar patterns and therefore the chaotic behavior is preserved. The maximum variation of mean firing rate (MFR) is about 0.5\% (Table \ref{tab:tab2}). This indicates that the type of the compressor has no significant impact on the network activity, even for the deepest possible compressor, i.e. [1024:1).

We use a 5 Hz sine wave as a teacher signal to train the network to adapt its dynamics. The training time is two seconds, then we simulate the network for one additional second to see its internal dynamics. Fig. \ref{fig:sine} shows the desired signal and the generated signal of each approximate network. As it can be seen, all approximate networks mimic the target signal accurately. As it is shown in Table \ref{tab:tab2}, the first four approximate networks show even lower mean squared error (MSE) than the exact network. Only the network with [1024:1) compressor has higher MSE, which is largely due to the time lag, not the shape of the generated signal. These interesting results are obtained while the maximum synaptic failure (Table \ref{tab:tab2}) is much less than what is in the Brain (see Section \ref{sec:deep}).

\begin{figure*}[tbp]
    \centering
	\includegraphics[width=\textwidth]{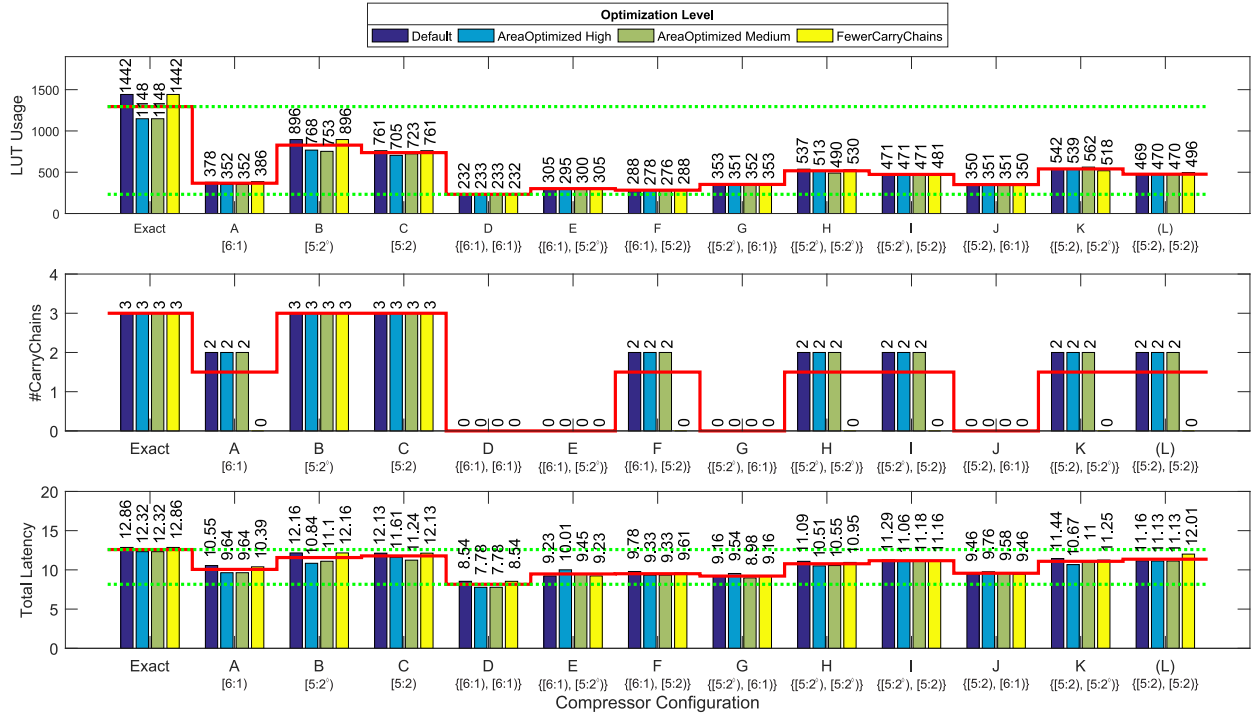}
	\caption{Resource consumptions and total latencies along with their mean, min, and max for different levels of optimizations. }
	\label{fig:resource}
\end{figure*}

\subsection{Hardware Synthesis Results}
To measure the latency and resource consumption improvement achieved through the use of the proposed architecture, we choose Xilinx Artix-7 as our target FPGA. First, we design all 12 approximate counters using Verilog HDL. Moreover, we have used the "dont\_touch" attribute to prevent optimizations for these counters. We next write a behavioral code to aggregate the outputs of the counters using a linear adder. Finally, we put the Vivado design suite to synthesize the RTL code automatically using its various optimization levels. In this work, we employ ``Default", ``AreaOptimized High", ``AreaOptimized Medium", and ``FewerCarryChains" optimization levels, which result in different resources usage and latencies.

Figure \ref{fig:resource} illustrates the resources usage including LUTs and CarryChains as well as the latency of the critical path in the nanosecond. All values are reported for four optimization levels mentioned above. For the sake of clarity, the means are shown in the red step graph, and the max and min values are represented with dashed green lines. Moreover, we show the mean values in Table \ref{tab:tab3}. In comparison with an exact counter, this table reveals that even the most accurate approximate counter (C) represents about 43\% reduction of the LUT consumption on average. This value is far higher for the (D) configuration where the LUT usage decreased by more than 82\%. From the latency point of view, while the best configuration, i.e. (C), shows only 6\% reduction, the latency of the configuration with the worst accuracy, i.e. (D), reduces by almost 35\%.

We have interpreted the two outputs of LUT5 in as positional and nonpositional systems. The synthesis results show that a [5:2$^{\diamondsuit}$) compressor causes more LUT usage when it is used as either one-level or second level of a two-level compressor. To verify this issue, we can compare configurations B, E, H, and K with C, F, I, and L respectively. Surprisingly, while the [5:2$^{\diamondsuit}$) compressor increases the LUT usage, it improves the latency. It is worth pointing out that, in this paper, we use a simple linear adder to aggregate the outputs of the approximate compressors. Using different methods, such as ones introduced in \cite{Parhami2009, Sklyarov2015a}, could adversely affect the trend.

\begin{table}[tbp]
    \centering
	\caption{Synthesis report for all configuration in comparison with exact one. all represented entries are mean values of four optimization levels.}
	\label{tab:tab3}
	\begin{tabular}{cccc}
		\hline
		Config.	&LUT Usage		&\#CarryChains	&Total Latency (ns)			\\
		\hline
		Exact	&\textbf{1295}	&3				&\textbf{12.59}\\
		(A) 	&367			&1.5			&10.06\\
		(B) 	&828			&3				&11.57\\
		(C) 	&737			&3				&11.78\\
		(D) 	&\textbf{232}	&0				&\textbf{8.16}\\
		(E)		&301			&0				& 9.48\\
		(F)		&282			&1.5			& 9.51\\
		(G)		&352			&0				& 9.21\\
		(H)		&517			&1.5			&10.78\\
		(I)		&473			&1.5			&11.17\\
		(J)		&350			&0				& 9.57\\
		(K)		&540			&1.5			&11.09\\
		(L)		&476			&1.5			&11.36\\
		\hline
	\end{tabular}
\end{table}

\section{Conclusion}
\label{sec4}
Inspired by synaptic transmission failure in the Brain, we have introduced an approximate architecture to calculate Hamming weight of a sparse binary vector. Our proposed method can be not only used to model the synaptic failure but also can be used as a general approximate Hamming weight calculator. This technique has split the Hamming weight calculator into two parts. First part is a LUT-based compressor that reduces the number of inputs of the second part, which is a smaller exact counter. This approach can improve both area and latency of FPGA based Hamming weight counters. Meanwhile, it could give a reasonable accuracy for sparse input vectors. Sparser vectors result in more accurate results. We proved this hypothesis using all 12 possible implementations of one-level and two-level compressors. Moreover, we applied both shallow and deep compressors to a network of 1024 spiking neurons. Then, we use this network to mimic a simple 5 Hz sine wave. Our simulation results show that even the deepest possible compressor could result in good learning performance while preserve the network chaotic activity. Finally, we show that all approximate implementations on FPGAs considerably outperform the exact one. 
\bibliographystyle{unsrt}  
\bibliography{references}

\begin{thebibliography}{10}

\bibitem{Merolla2014b}
P~A Merolla, J~V Arthur, R~Alvarez-Icaza, A~S Cassidy, J~Sawada, F~Akopyan, B~L
  Jackson, N~Imam, C~Guo, Y~Nakamura, B~Brezzo, I~Vo, S~K Esser, R~Appuswamy,
  B~Taba, A~Amir, M~D Flickner, W~P Risk, R~Manohar, and D~S Modha.
\newblock {A million spiking-neuron integrated circuit with a scalable
  communication network and interface}.
\newblock {\em Science}, 345(6197):668--673, aug 2014.

\bibitem{Davies2018}
Mike Davies, Narayan Srinivasa, Tsung-Han Lin, Gautham Chinya, Yongqiang Cao,
  Sri~Harsha Choday, Georgios Dimou, Prasad Joshi, Nabil Imam, Shweta Jain,
  Yuyun Liao, Chit-Kwan Lin, Andrew Lines, Ruokun Liu, Deepak Mathaikutty,
  Steven McCoy, Arnab Paul, Jonathan Tse, Guruguhanathan Venkataramanan,
  Yi-Hsin Weng, Andreas Wild, Yoonseok Yang, and Hong Wang.
\newblock {Loihi: A Neuromorphic Manycore Processor with On-Chip Learning}.
\newblock {\em IEEE Micro}, 38(1):82--99, jan 2018.

\bibitem{Jaeger2001}
Herbert Jaeger.
\newblock {The "echo state" approach to analysing and training recurrent neural
  networks}.
\newblock {\em GMD Report 148}, 2001.

\bibitem{Maass2002}
Wolfgang Maass, Thomas Natschl{\"{a}}ger, and Henry Markram.
\newblock {Real-Time Computing Without Stable States: A New Framework for
  Neural Computation Based on Perturbations}.
\newblock {\em Neural Computation}, 14(11):2531--2560, nov 2002.

\bibitem{Sussillo2009}
David Sussillo and L.F. Abbott.
\newblock {Generating Coherent Patterns of Activity from Chaotic Neural
  Networks}.
\newblock {\em Neuron}, 63(4):544--557, aug 2009.

\bibitem{Nicola2017}
Wilten Nicola and Claudia Clopath.
\newblock {Supervised learning in spiking neural networks with FORCE training}.
\newblock {\em Nature Communications}, 8(1):2208, dec 2017.

\bibitem{Gilra2017}
Aditya Gilra and Wulfram Gerstner.
\newblock {Predicting non-linear dynamics by stable local learning in a
  recurrent spiking neural network}.
\newblock {\em eLife}, 6, nov 2017.

\bibitem{Gao2016}
Junling Gao, Jicong Fan, Bonnie Wai~Yan Wu, Zhiguo Zhang, Chunqi Chang,
  Yeung-Sam Hung, Peter Chin~Wan Fung, and Hin hung Sik.
\newblock {Entrainment of chaotic activities in brain and heart during MBSR
  mindfulness training}.
\newblock {\em Neuroscience Letters}, 616:218--223, mar 2016.

\bibitem{Kaveh2018}
Kaveh Akbarzadeh-Sherbaf, Behrooz Abdoli, Saeed Safari, and Abdol-Hossein
  Vahabie.
\newblock {A Scalable FPGA Architecture for Randomly Connected Networks of
  Hodgkin-Huxley Neurons}.
\newblock {\em Frontiers in Neuroscience}, 12, oct 2018.

\bibitem{Seyed-allaei2015}
Hamed Seyed-allaei.
\newblock {Phase diagram of spiking neural networks}.
\newblock {\em Frontiers in Computational Neuroscience}, 9, mar 2015.

\bibitem{Hessler1993}
Neal~A. Hessler, Aneil~M. Shirke, and Roberto Malinow.
\newblock {The probability of transmitter release at a mammalian central
  synapse}.
\newblock {\em Nature}, 366(6455):569--572, 1993.

\bibitem{Markram1996}
H.~Markram and M.~Tsodyks.
\newblock {Redistribution of synaptic efficacy between neocortical pyramidal
  neurons}.
\newblock {\em Nature}, 382(6594):807--810, 1996.

\bibitem{Kurzak2010}
Hari Subramoni, Fabrizio Petrini, Virat Agarwal, , and Davide Pasetto.
\newblock High performance topology-aware communication in multicore
  processors.
\newblock In Jakub Kurzak, David~A. Bader, and Jack Dongarra, editors, {\em
  Scientific Computing with Multicore and Accelerators}, pages 443--459. CRC
  Press, NY, USA, 1st edition, 2010.

\bibitem{Morancho2014}
Enric Morancho.
\newblock {A hybrid implementation of Hamming weight}.
\newblock In {\em 2014 22nd Euromicro International Conference on Parallel,
  Distributed, and Network-Based Processing}, pages 84--92. IEEE, feb 2014.

\bibitem{Parhami2009}
Behrooz Parhami.
\newblock {Efficient Hamming weight comparators for binary vectors based on
  accumulative and up/down parallel counters}.
\newblock {\em IEEE Transactions on Circuits and Systems II: Express Briefs},
  56(2):167--171, feb 2009.

\bibitem{Sklyarov2015a}
Valery Sklyarov and Iouliia Skliarova.
\newblock {Design and implementation of counting networks}.
\newblock {\em Computing}, 97(6):557--577, jun 2015.

\bibitem{Pedroni2004}
V~A Pedroni.
\newblock {Compact Hamming-comparator-based rank order filter for digital VLSI
  and FPGA implementations}.
\newblock {\em Proceedings of the 2004 IEEE International Symposium on Circuits
  and Systems (ISCAS'2004)}, 1:585--588, 2004.

\bibitem{Swartzlander1973}
Earl~E. Swartzlander.
\newblock {Parallel Counters}.
\newblock {\em IEEE Transactions on Computers}, C-22(11):1021--1024, nov 1973.

\bibitem{Parhami2010}
Behrooz Parhami.
\newblock {\em {Computer arithmetic : algorithms and hardware designs}}.
\newblock Oxford University Press, NY, USA, 2nd edition, 2010.

\end{thebibliography}


% George Kour and Raid Saabne.
% \newblock Real-time segmentation of on-line handwritten arabic script.
% \newblock In {\em Frontiers in Handwriting Recognition (ICFHR), 2014 14th
%   International Conference on}, pages 417--422. IEEE, 2014.

% George Kour and Raid Saabne.
% \newblock Fast classification of handwritten on-line arabic characters.
% \newblock In {\em Soft Computing and Pattern Recognition (SoCPaR), 2014 6th
%   International Conference of}, pages 312--318. IEEE, 2014.

% \bibitem{hadash2018estimate}
% Guy Hadash, Einat Kermany, Boaz Carmeli, Ofer Lavi, George Kour, and Alon
% \newblock Estimate and replace: A novel approach to integrating deep neural
%   networks with existing applications.
% \newblock {\em arXiv preprint arXiv:1804.09028}, 2018.

% \end{thebibliography}

\end{document}